\documentclass{article}
\usepackage{spconf}
\usepackage{fancyhdr}
\usepackage{amsmath,graphicx}
\usepackage{amsfonts}
\usepackage{amssymb}

\usepackage{color}

\usepackage{algorithm}
\usepackage{algorithmic}
\usepackage[pagebackref=true,breaklinks=true,letterpaper=true,colorlinks,bookmarks=false]{hyperref}
\usepackage{multirow}
\usepackage{booktabs}
\usepackage{makecell}
\usepackage{array}

\pdfoutput=1
    \title{A two-step learning method for detecting landmarks \\on faces from different domains}
\name{Bruna Vieira Frade \hspace{2cm} Erickson R. Nascimento}
\address{Universidade Federal de Minas Gerais (UFMG), Brazil \\\{brunafrade, erickson\}@dcc.ufmg.br}
\begin{document}
%
\maketitle

\thispagestyle{fancy}
\fancyhf{}
\fancyfoot[L]{978-1-4799-7061-2/18/\$31.00 \copyright2018 IEEE}
\chead{In Proceedings of the 2018 IEEE International Conference on Image Processing (ICIP) \\ The final publication is available at: http://dx.doi.org/10.1109/ICIP.2018.8451026}
\setlength{\headsep}{0.08 in}

\begin{abstract}
The detection of fiducial points on faces has significantly been favored by the rapid progress in the field of machine learning, in particular in the convolution networks. However, the accuracy of most of the detectors strongly depends on an enormous amount of annotated data. In this work, we present a domain adaptation approach based on a two-step learning to detect fiducial points on human and animal faces. We evaluate our method on three different datasets composed of different animal faces (cats, dogs, and horses). The experiments show that our method performs better than state of the art and can use few annotated data to leverage the detection of landmarks reducing the demand for large volume of annotated data.
\end{abstract}
\begin{keywords}
	Landmarks detection, Machine Learning, Domain Adaptation,  Human faces, Animal faces
\end{keywords}

\section{Introduction}
\label{sec:intro}

Detecting landmarks embedded with semantic information from images is one of the key challenges in image processing and computer vision fields. In general, landmarks or fiducial points are related to discriminative locations in the image, frequently embedding some meaning. For example, in human or animal faces, a landmark locates regions comprising the eyes, eyebrows, mouth, and the tip of the nose. After all, the automatic estimation of landmarks on faces has a myriad of applications such as faces recognition, game animation, avatars, and transferring facial expressions~\cite{sumner2004}. 

Despite remarkable advances in detecting landmarks on human faces, most of the methods require a large number of annotated data. For every different type of face like an animal such as a cat or a dog face, we still have to use the time-consuming process of annotating each landmark for a considerable amount of data. In other words, although detecting the same type of landmarks  present in a large dataset of human faces (\emph{e.g.}, eyes, nose, \emph{etc}.), we need to build an entirely new dataset. Thus, a central challenge in facial landmark detection is how to use the annotation available in big datasets such those for human faces to improve the detection of similar landmarks but on different types of faces.

In this work, we present a domain adaptation algorithm based on deep learning to detect landmarks on human and non-humans faces. Our method builds a landmark detector by performing two tasks: i) learning to identify landmarks in a supervised way by using labeled data of human faces (source domain) and ii) learning to reconstruct non-human faces with unlabeled data (target domain). The final representation of our method preserves the discriminability from the labeled data and encodes the landmarks locations of the target domain. This capability suggests that the performance of our method stems from the creation of a single representation that encodes the structure information of non-human face and relevant features for the landmark detection on the human face.

According to our experiments, our method outperformed the state-of-the-art landmarks detector for interspecies face
~\cite{interspecies} up to $10\%$ precision gap. We evaluated our method on a variety of type of faces, where it was capable of learning the cross-domain landmark detection task, without requiring a big collection of annotated data.

\paragraph*{Related Work.}

In the past several years, a popular approach for detecting fiducial points was based on using classifiers ~\cite{adaboost,arvore,belhumeur2013localizing}. However, recently we have witnessed an explosion of approaches to learning features found on convolution neural networks. Sun et al.~\cite{sun2013deep} used three levels of convolutional neural networks (CNN) to estimate the position of landmarks on faces. Thanks to the high-level global characteristics extracted from the entire face, their method increases the precision of the landmarks detection. To minimize occlusion effects, Zhang et al.~\cite{zhang2014facial} proposed a multitask learning to optimize landmark detection through heterogeneous but correlated tasks, \emph{i.e.}, head pose and facial attributes inference.  Zhang et al.~\cite{CFAN} aim at refining the alignment in each stage. Using an autoenconder approach, they tried to predict landmarks quickly through low-quality images and progressively improving previous results while increasing its resolution. The work of Yu et al.~\cite{eccv16} used a cascading approach called deep deformation network (DDN). The DDN learns to extract the shape information and then uses a landmark transformation network to estimate the local parameterized deformation aiming to refine first step results.

\begin{figure*}[t!]
	\centering
	\includegraphics[width=0.9\textwidth]{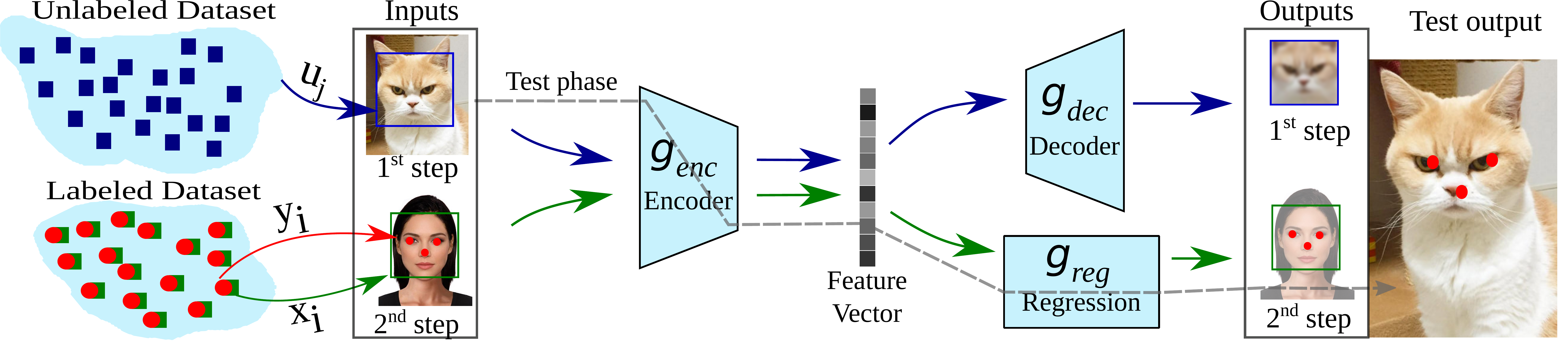}
	\caption{Illustration of our two-step learning with the supervised (regression) and the unsupervised steps. In the first step, we encode the face features of a unlabeled image $\mathbf{u_i}$ using $g_{enc}$ network. Then we apply a decode function to reconstruct the input, \emph{i.e.}, $g_{dec}(g_{enc}(\mathbf{u}_i))$. In the second step, we feed a regression function $ g_{reg}$ that learns the coordinates of each landmark by estimating the error between the ground truth $ \mathbf{y_i}$ and the prediction $g_{reg}(g_{enc}(\mathbf{x_i}))$.}
	\label{fig:drkdn}
\end{figure*}

A significant disadvantage of most CNN based methods is the need for large-scale datasets. Yang et al.~\cite{sheep} and Rashid et al.~\cite{interspecies} propose to adjust the landmark learning from human faces to other target topology. The work of Rashid et al. searches for similar human faces for each animal sample in an unsupervised method. Yang et al. interpolate face characteristics by cascade regression aiming at detecting through shape using fewer samples data of the target domain. A recent approach is the Deep Reconstruction Classification Network (DRCN)~\cite{ghifary2016deep}. The DRCN jointly learns a shared encoding representation for the digits classification task. Inspired by the DRCN approach, our method is also based on a two-step learning domain adaptation approach, but different from it, we used the learned encoding for a regression problem solution in the faces domain.

\section{Methodology}

Let $\mathcal{D}^{source}$ be a large set of labeled images like human faces dataset, and $\mathcal{D}^{target}$ be a set with a small number of labeled samples, \emph{e.g.}, dog's face. Our methodology has been designed to detect landmarks in both $\mathcal{D}^{source}$ and $\mathcal{D}^{target}$ domains.  Our formulation is based on a two-step learning approach, wherein the first step it learns to reconstruct images from $\mathcal{D}^{target}$ using an unsupervised strategy and in the second step it solves a regression problem in a supervised way predicting the coordinates of the landmarks in faces from $\mathcal{D}^{source}$. Figure~\ref{fig:drkdn} illustrates the whole process.

%
%



The reconstruction and detection steps rely on the network $g_{enc}(\cdot)$ that learns to encode discriminative features of faces on different domains and tasks. This function plays a key role in building a model capable of reconstructing faces and detecting landmarks.

\paragraph*{Reconstruction.} Let $\mathbf{u}_i \in \mathcal{D}^{target}$ be the $i$-th unlabeled image of an animal. After encoding the face features using $g_{enc}(\mathbf{u_i})$, we apply a decode function $g_{dec}(\cdot)$ in order to create an output $g_{dec}(g_{enc}(\mathbf{u}_i))$ as close as possible to $\mathbf{u}_i$. 

In other words, we train the network to minimize the loss function: 
\begin{equation}
\mathcal{L}_{rec} = \frac{1}{n} \sum _{i=1}^{n} \| \mathbf{u}_i - g_{dec}(g_{enc}(\mathbf{u}_i)) \|^2, \label{eq:1}  
\end{equation}
\noindent where $n$ is the number of images in $\mathcal{D}^{target}$ with no annotations, because this step does not require labeled data.

\paragraph*{Landmark detection.} In the second step, we apply a supervised approach to learn to detect landmarks on faces using annotated data. Let $\mathbf{x}_i \in \mathcal{D}^{source}$ be the i-$th$ image of a person and $\mathbf{y}_i \in \mathbb{R}^6$ be the image coordinates of the landmarks, \emph{i.e.}, eyes and nose.

\begin{algorithm}[t!]
	\caption{Two-step learning for landmark detection.}
	\label{alg:two-step-learning}
	\begin{algorithmic}[1]
		\STATE Labeled dataset: $\mathcal{D}^{source}=\{(\mathbf{x}_{i},\mathbf{y}_{i})\}_{i=1}^{m}$
		\STATE Unlabeled dataset:  $\mathcal{D}^{target}=\{\mathbf{u}_j\}_{j=1}^{n}$
		
		\FOR {\textbf{each}  $e < totalEpoch$}
		\FOR {\textbf{each} $\text{batch}_{t}\in \mathcal{D}^{target}$ and $\text{batch}_{s}\in\mathcal{D}^{source}$}
		\STATE ForwardUnsupervised($\text{batch}_{t}$) 
		\STATE ComputeReconstructionError: $\mathcal{L}_{rec}(\text{batch}_{t})$
		\STATE UpdateWeights of $g_{dec}$ and $g_{enc}$ 
		\STATE ForwardSupervised($\text{batch}_{s}$) 
		\STATE ComputeRegressionError: $\mathcal{L}_{reg}(\text{batch}_{s})$
		\STATE UpdateWeights of $g_{reg}$ and $g_{enc}$ 
		\ENDFOR
		
		\ENDFOR
	\end{algorithmic}
\end{algorithm}    
In our approach, the supervised branch of our architecture uses the representation learned in the unsupervised step  for feeding the supervised the regression function $ g_{reg}(\cdot)$. The regression function learns the coordinates of each landmark by computing the error between the ground truth $ \mathbf{y_i}$ and the prediction $g_{reg}(g_{enc}(\mathbf{x_i}))$ as:
\begin{equation}
\mathcal{L}_{reg} = \frac{1}{m} \sum _{i=1}^{m} \text{MAE}\left( \mathbf{y}_i - g_{reg}(g_{enc}(\mathbf{x}_i)) \right),
\label{eq:2} 
\end{equation}
\noindent where $m$ is the size of the annotated set and $\text{MAE}$ is the mean absolute error between two vectors of size $k$, \emph{i.e.}, $\text{MAE}(\mathbf{a},\mathbf{b}) = \frac{1}{k} \sum _{i=1}^{k} \vert \mathbf{a}_i - \mathbf{b}_i \vert$.

\begin{figure*}[t!]
	\centering
	\begin{tabular}{ccc}
		\includegraphics[width=0.34\textwidth]{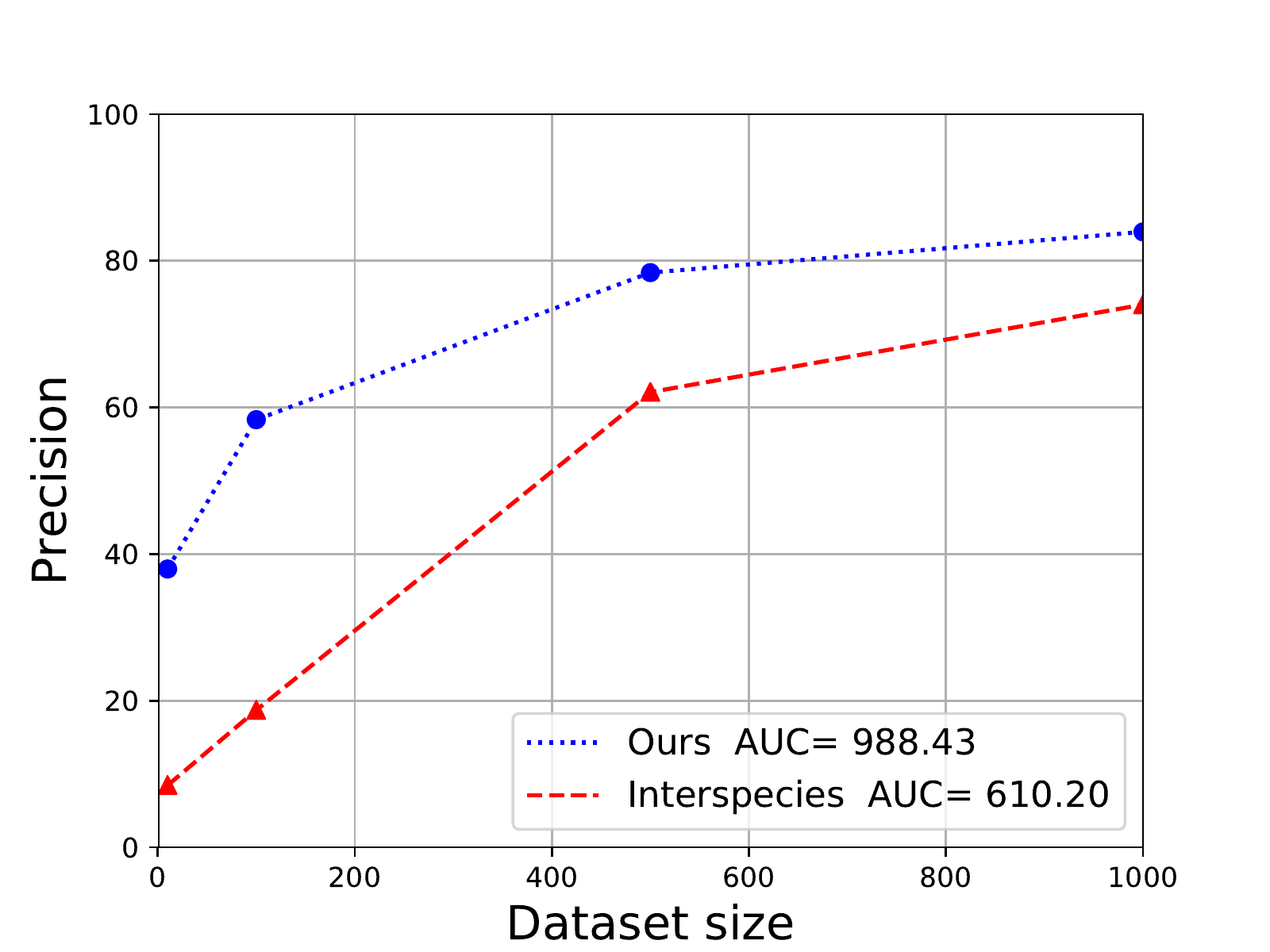}
		&
		\includegraphics[width=0.32\textwidth]{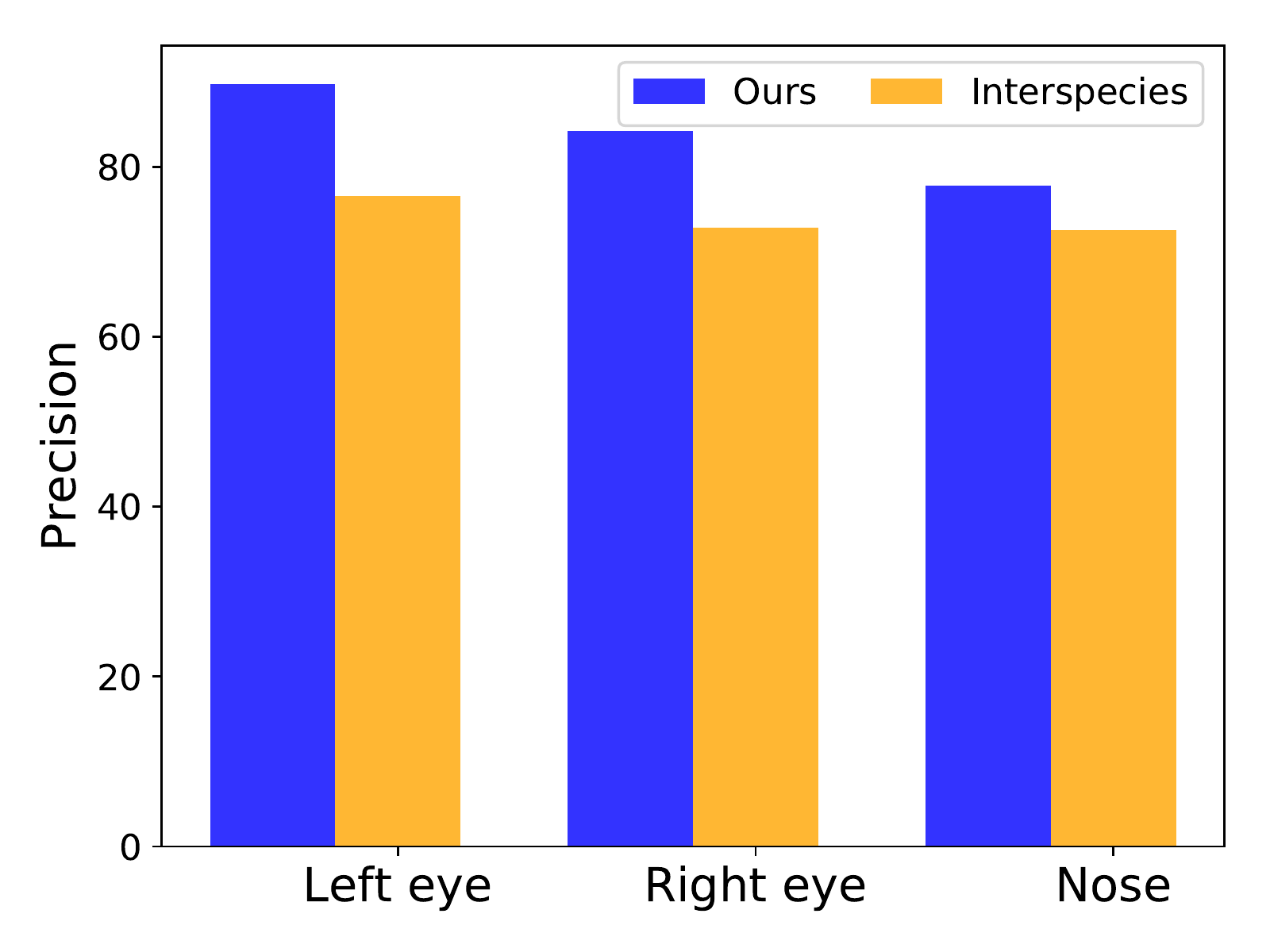}     &
		\includegraphics[width=0.23\textwidth]{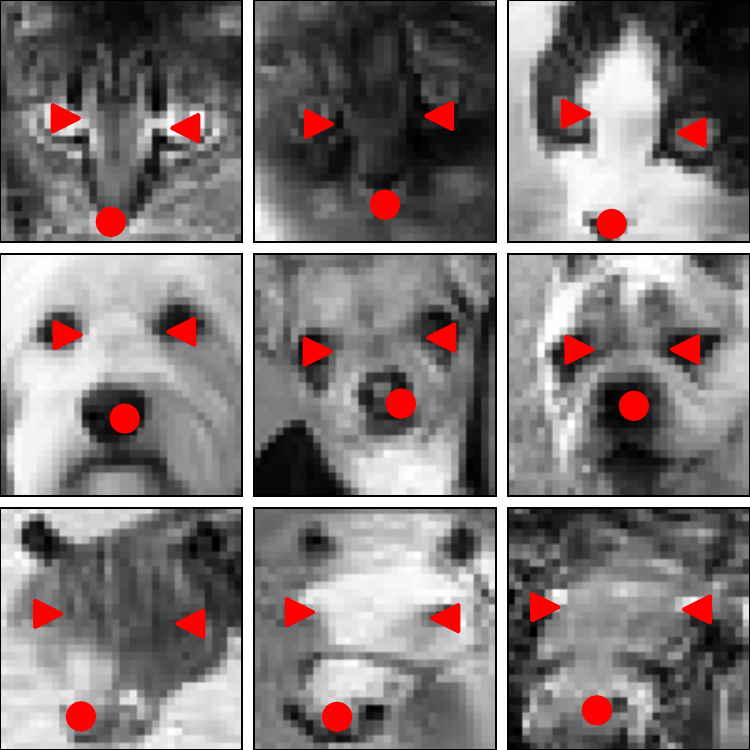} \\
		(a) & (b) & (c) 
	\end{tabular}
	\caption{(a) ROC curve of our method and the Interspecies using different numbers of labeled images of the target domain (from $10$ to $1{,}000$ images). Larger AUC means better performance. (b) Precision  of the predicted landmarks using with $1{,}000$ of labeled images. (c) Qualitative results for cats, dogs, and horses training our method with $100$ of labeled images.}
	\label{fig:SOTAcomparison}
\end{figure*}

In our two-step learning, we address the lack of annotated data in the target domain. Algorithm~\ref{alg:two-step-learning} depicts the learning loop. Instead of starting the learning from random weights, we use the patterns learned from a reconstruction task. These patterns help to leverage the landmark detection when solving the regression in a supervised way but in a different domain. Thanks to this strategy, we can extract robust features by hallucinating face features of datasets with a few annotations. Our hypothesis is grounded in the idea that we can learn a function that maps similar features in both domains. Thus, with this mapping, we can simultaneously perform the regression and reconstruction tasks and simultaneously learn how to detect landmarks in the target domain.

Unlike DRCN~\cite{ghifary2016deep} that updates the weights of a classification net using all labeled batches and then adjusts the encoder weights in the reconstruction with all unlabeled batches, our approach updates the weights of the reconstruction and regression nets for each batch intercalating both steps. This fact plays a key role in the development of a robust and flexible strategy that can work with unlabeled data in the target domain or a small collection of annotations.

\section{Experiments}


\paragraph*{Datasets.} In our experiments, the labeled data are from human faces of the Keggle~\cite{keggle}. For the target domain, \emph{i.e.}, animal faces, aside from  dataset of cat faces~\cite{catdataset}, we also evaluated our approach on dogs faces~\cite{KhoslaYaoJayadevaprakashFeiFei_FGVC2011} and horse faces~\cite{interspecies}. All datasets have the faces annotated with $3$ landmarks: left and right eyes, and nose. We used the Euclidean distance as  evaluation metric for the location predictions and margin of error of $10\%$ of the image size (in the dataset we used in the experiments correspondes to a radius of $3$ pixels) to classify a detection as correct. We performed data augmentation in source and target datasets by applying a random rotations in the images from $ -30 $ to $ 30 $ degrees. Moreover, we applied translations and noise in the target set to ensure robust results in the reconstruction step.

\paragraph*{Baselines.} We compared our method with a standard convolutional network (ConvNet) for supervised landmark detection. This ConvNet has the same architecture of our supervised net and it was trained on the target domain with labeled data. We also pit our detector against the state-of-the-art method on landmark detection for faces in different topologies called Interspecies~\cite{interspecies}. We used the code provided by the authors with our configuration of training and test data ranging from $10$ to $1{,}000$ images. 
\\\\
\textbf{Implementation}. The encoder $g_{enc}(\cdot)$ has five convolution layers with $3\times3$ filters and padding $1\times1$: $300$ filters in conv1, $250$ filters in conv2, $200$ filters in conv3, $150$ filters in conv4, and conv5 with $100$ filters; two pooling layers $2\times2$ after the first and second convolutional layers (pool1 and pool2) and a fully connected layer (fc4) with $500$ neurons. The decoder net $g_{enc}$ is the mirror image of the encoder architecture. In the regression, we feed a fully connected layer (fc-regressor) with the output from the $g_{enc}$ net. We used ReLU in all hidden and output layers and hyperbolic tangent activation for the regression layer. We used a learning rate equals to $3\times10^{-4}$ for point detection and the reconstruction. In the training, we ran $500$ epochs and used a batch size equals to $128$. The size of all input images is $32\times32$. All source code and experimental data will be publicly available.
 
\begin{figure*}[!t]
	\centering
	\begin{tabular}{ccc}
		\includegraphics[width=0.32\textwidth]{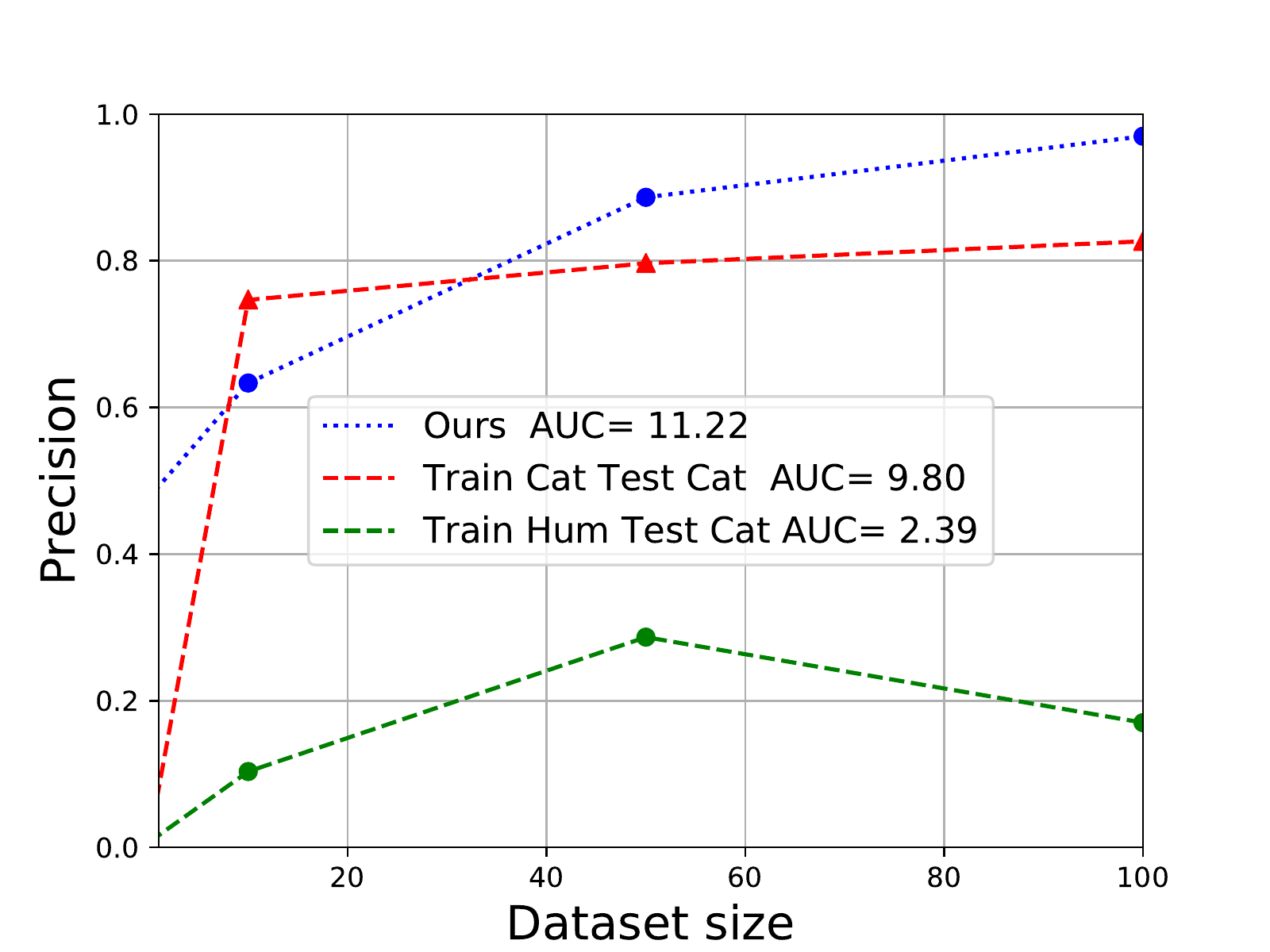}
		&
		\includegraphics[width=0.32\textwidth]{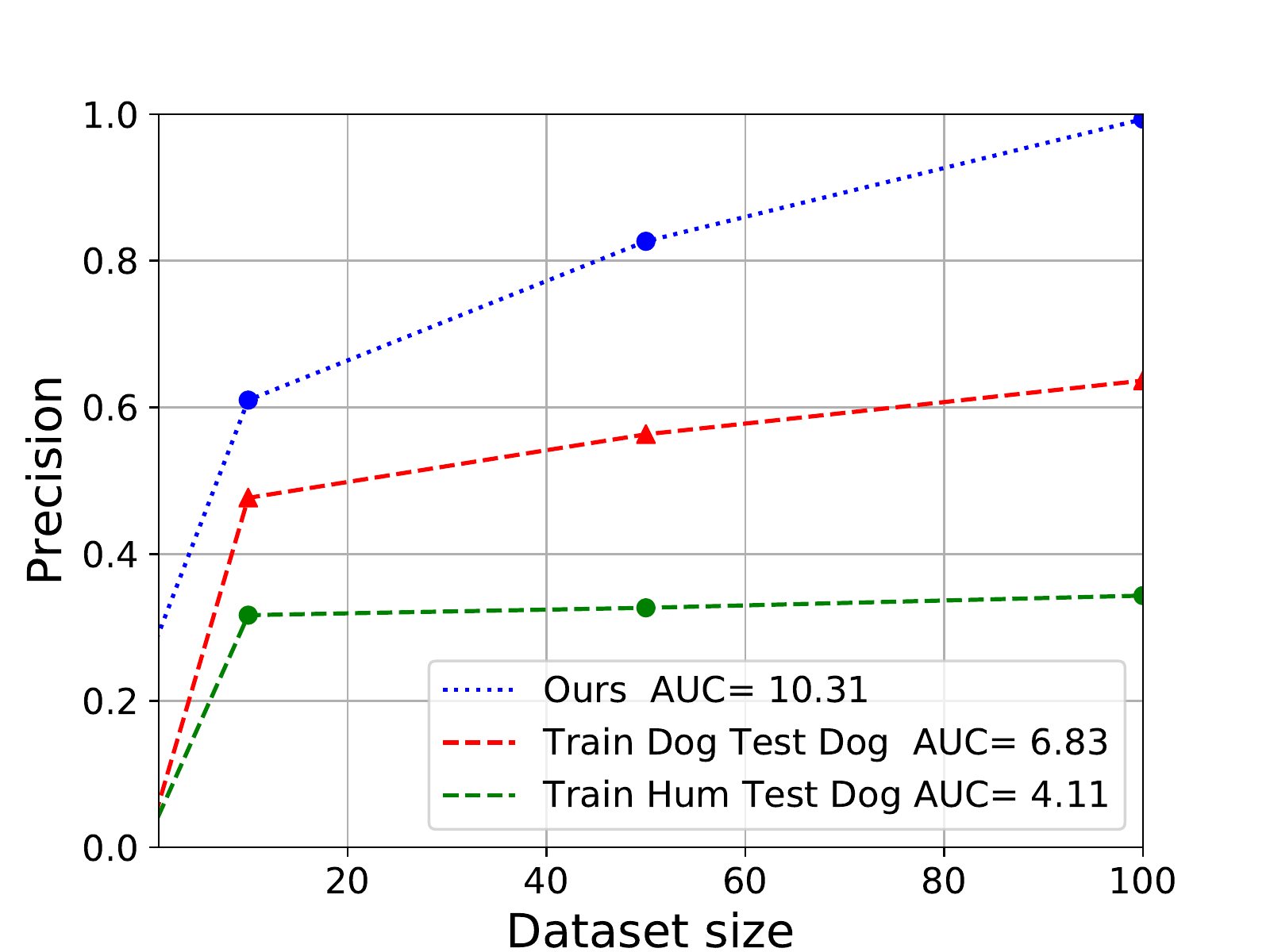}
		&
		\includegraphics[width=0.32\textwidth]{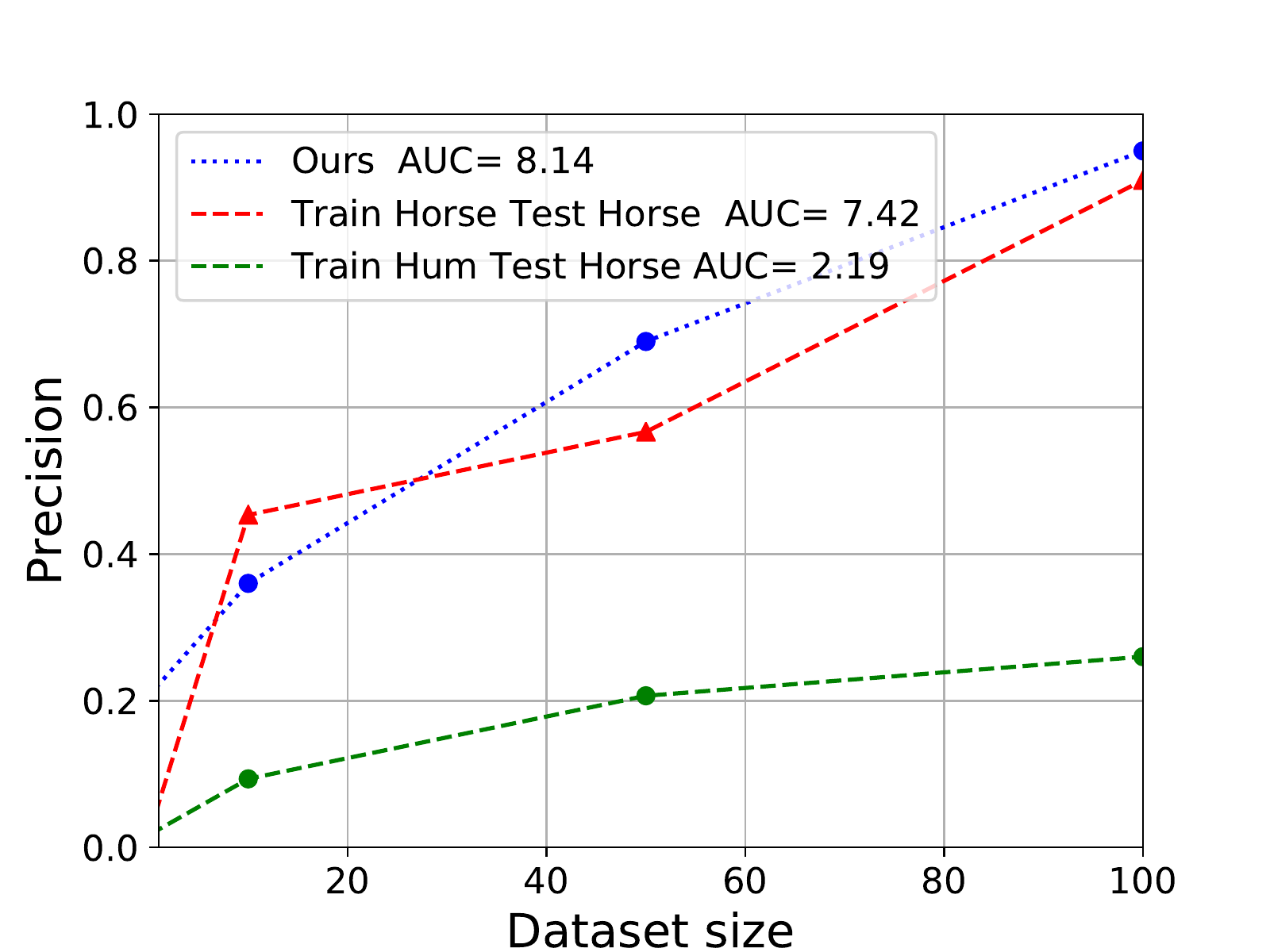} \\
		(a) \textit{CAT} & (b) \textit{DOG} & (c) \textit{HORSE}
		
	\end{tabular}    
	\caption{ROC curve varying the number of labeled data for CAT, DOG and HORSE datates. Our method (blue curve) was superior than ConvNet trained with samples from target domain (red curve) and source domain (green curve).}
	\label{fig:usinglabeleddatatarget}
\end{figure*}

\subsection{Comparison against the state of the art}

We compared our work with the Interspecies method proposed by Rashid et al.~\cite{interspecies}, the current state of the art in detecting landmarks on animal faces. In the experiments, we varied the size of training from $10$ to $1{,}000$ images.

Figure~\ref{fig:SOTAcomparison} (a) shows the Receiver Operating Characteristic (ROC) of our method and the Interspecies. On can clearly see that our method outperformed the Interspecies method. Our method obtained $ 83\% $ of the precision while the Interspecies method obtained  $ 73.2\% $ of correct detections. We also evaluate prediction precision for each landmark. As one can see in the bars of Figure~\ref{fig:SOTAcomparison} (b), our method has larger precision for all the landmarks, \emph{i.e.}, nose and eyes locations. 

From these results, we can draw the following observations. First, even when there is no labeled data from the target domain, our method performs better than Interspecies. The experiments show that our approach can  hallucinate features by adapting learnt features in a reconstruction task from a different domain. Second, whenever available, our method can use few annotated data to leverage the detection of landmarks reducing the demand of large volume of annotated data.

\subsection{Ablation analysis}

For a more detailed performance assessment, we also evaluate our method in two experiments: in the first experiment we used only unlabeled data from the target domain and labeled data from the source domain; in the second experiment, we gradually increased the size of labeled dataset from the target domain starting from $5$ up to $100$ images. 
\begin{table}[t!]
	\centering
		\caption{Area Under the Curve (AUC) for ROC of precision for our method and a ConvNet trained with labeled data from a human faces dataset. Best in bold.}
		\begin{tabular}{@{}lccc@{}}
			\bottomrule
			\multirow{2}{*}{\thead{\textbf{Method}}} & \multicolumn{3}{c}{\thead{\textbf{Dataset (AUC)}}}  \\ 
			\cmidrule(lr){2-4}	
			& \thead{\textit{CAT}}	& \thead{\textit{DOG}}	&   \thead{\textit{HORSE}}   \\	\midrule
			\textsc{Ours}	&   \textbf{86.3}	&   \textbf{82.3}	&   \textbf{79.97}	\\
			\textsc{ConvNet}	&   77.33	&           81.10	&   76.65	\\ 
			\bottomrule
		\end{tabular}
 
	\label{tab:nolabeltargetdomain}

\end{table}
\\
\textbf{Only unlabeled data from the target domain.} Table~\ref{tab:nolabeltargetdomain} shows the area under the curve of the ROC curves for our model trained with no labeled data from target domain and a ConvNet trained with source domain only. When comparing with ConvNet, our method improved the detection in all three datasets. Together these results show that the hypothesis of leveraging feature from one domain to another holds for detecting landmarks on faces.
\\
\textbf{Using a few of labeled data from the target domain.} In this experiment, we use some labeled data from the target domain in the regression step. The idea is to analyze the performance of our approach when using a small number of labeled data from the target domain. For each batch, we forward a subset of labeled data from the target domain and solve the regression according to the error between the prediction and the ground truth of the target domain. Then, the network weights are updated. We used four different number of labeled images: $0,10,50$, and $100$. Figure~\ref{fig:usinglabeleddatatarget} shows the precision for the three datasets. It is noteworthy the rapid increase in the precision of our method and the superior performance (larger AUC). This fact can be explained by the transfer learning from the source domain and the features learning in the facial reconstruction in the target domain.

\section{Conclusion}
We presented a novel method for detecting landmarks on faces in different domains such as human and animal faces. Our method is based on a two-step learning (supervised and unsupervised) that reduces the need for a large annotated data. The experiments show that our method performed better than the state-of-the-art method even when there is a small collection of annotated data from the target domain.
\section{ACKNOWLEDGMENTS}
The authors would like to thank the agencies CAPES, CNPq, FAPEMIG, Vale Institute of Technology (ITV), and Petrobras for funding different parts of this work.

\bibliographystyle{IEEEbib}

\bibliography{refs}

\end{document}